\documentclass[sigconf]{acmart}

\AtBeginDocument{%
  \providecommand\BibTeX{{%
    \normalfont B\kern-0.5em{\scshape i\kern-0.25em b}\kern-0.8em\TeX}}}

\copyrightyear{2021}
\acmYear{2021} 
\setcopyright{iw3c2w3}
\acmConference[WWW '21]{Proceedings of the Web Conference 2021}{April 19--23, 2021}{Ljubljana, Slovenia} 
\acmBooktitle{Proceedings of the Web Conference 2021 (WWW '21), April 19--23, 2021, Ljubljana, Slovenia}
\acmPrice{}
\acmDOI{10.1145/3442381.3449806}
\acmISBN{978-1-4503-8312-7/21/04}
\usepackage[utf8]{inputenc} 
\usepackage[T1]{fontenc}    
\usepackage{hyperref}       
\usepackage{url}            
\usepackage{booktabs}       
\usepackage{sidecap}        
\usepackage{multirow}       
\usepackage{color}
\usepackage{graphicx}
\definecolor{LightCyan}{rgb}{0.9,0.9,0.9}
\usepackage{amsfonts}       
\usepackage{nicefrac}       
\usepackage{microtype}      

\def\ourdataset{TiMoS}
\def\ourdatasetfull{Tropes in Movie Synopses}
\begin{document}


\title{Situation and Behavior Understanding by Trope Detection on Films}
\author{
Chen-Hsi Chang$^{1*}$, Hung-Ting Su$^{1*}$, Jui-Heng Hsu$^{1}$, Yu-Siang Wang$^{2}$, Yu-Cheng Chang$^{1}$, \mbox{Zhe Yu Liu$^{1}$}, Ya-Liang Chang$^{1}$, Wen-Feng Cheng$^{1,3}$, Ke-Jyun Wang$^{1}$, and Winston H. Hsu$^{1}$
}
\thanks{$^*$ Equal contribution.}
\affiliation{
\institution{$^{1}$National Taiwan University, $^{2}$University of Toronto, $^{3}$Microsoft}
}
\renewcommand{\shortauthors}{Chang, Su, Hsu, Wang, Chang, Liu, Chang, Cheng, Wang, and Hsu}

\begin{abstract}

The human ability of deep cognitive skills is crucial for the development of various real-world applications that process diverse and  abundant user generated input. 
While recent progress of deep learning and natural language processing have enabled learning system to reach human performance on some benchmarks requiring shallow semantics, such human ability still remains challenging for even modern contextual embedding models, as pointed out by many recent studies \cite{bertnot-ettinger-2020-bert,sysb-linzen-2020-accelerate,sysb-sinha-etal-2019-clutrr,sys-DBLP:journals/corr/abs-1911-02969,definerc-dunietz-etal-2020-test}. Existing machine comprehension datasets assume sentence-level input, lack of casual or motivational inferences, or can be answered with question-answer bias. Here, we present a challenging novel task, \textbf{trope detection} on films, in an effort to create a situation and behavior understanding for machines. Tropes are frequently used storytelling devices for creative works. Comparing to existing movie tag prediction tasks, tropes are more sophisticated as they can vary widely, from a moral concept to a series of circumstances, and embedded with motivations and cause-and-effects. 
We introduce a new dataset, \textbf{Tropes in Movie Synopses (TiMoS)}, with 5623 movie synopses and 95 different tropes collecting from a Wikipedia-style database, TVTropes. 
We present a multi-stream comprehension network (MulCom) leveraging multi-level attention of words, sentences, and role relations. 
  Experimental result demonstrates that modern models including BERT contextual embedding, movie tag prediction systems, and relational networks, perform at most 37\% of human performance (23.97/64.87) 
  in terms of F1 score. Our MulCom outperforms all modern baselines, by 1.5 to 5.0 F1 score and 1.5 to 3.0 mean of average precision (mAP) score. We also provide a detailed analysis and human evaluation to pave ways for future research. 
\end{abstract}

\maketitle

\section{Introduction}

\begin{figure*}
    \centering
    \includegraphics[width=\textwidth]{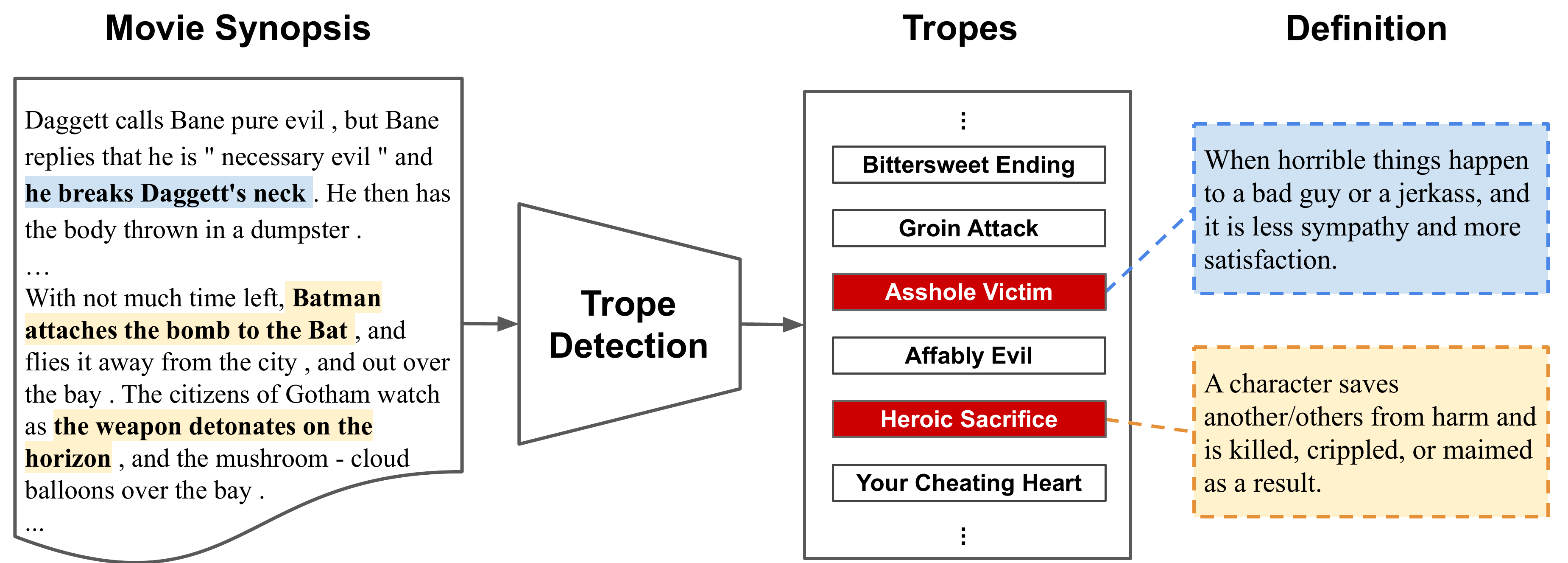}
    \caption{An example of \textbf{Trope Detection on Films}. Detecting the tropes requires deep cognitive skills of a learning system. For example, we could see a \textit{victim} (Daggett in the first paragraph) is killed, but to determine if it is an \textit{Asshole Victim}, one needs to understand what the victim did beforehand, which shifts sympathy away from him. Similarly, in the second paragraph, an agent might notice that Batman flies away from the city with shallow semantics portrayed. However, one needs to comprehend the motivation of the left (to save the city) and the causality of the mushroom to reason that it is a \textit{Heroic Sacrifice}. 
    }
    \label{fig:tropedet_task}
\end{figure*}

Human cognitive abilities to understand and process abundant, daily, diverse, user-generated language content are crucial for various applications that face or utilize human written text like recommendation systems, chat-bots, question answering, personal assistant, etc. Some modern models have achieved human-level performance on benchmarks with \textit{shallow semantics} on texts and images, such as ImageNet image classification \cite{imagenet_cvpr09}, DBPedia text classification \cite{dbpedia-swj}, or SQuAD reading comprehension \cite{squad-rajpurkar-etal-2016-squad}. However, \textit{deep cognitive skills}\footnote{Deep cognitive skills include but not limited to consciousness, systematic generalization, and casual and motivational comprehension \cite{bengio2019system}. This work mainly focuses on above three.}, including consciousness, systematic generalization, causality, and comprehension, remain difficult for learning systems, as highlighted by others \cite{bengio2019system}. Contextual embedding models \cite{elmo-Peters:2018,devlin-etal-2019-bert,xlnet-NIPS2019_8812}, particularly, perform great on many datasets with sufficient parameters and unsupervised pre-training data. Nevertheless, recent research investigated the drawback of contextual embedding, including but not limited to (1) being insensitive to commonsense and role-based events \cite{bertnot-ettinger-2020-bert}, (2) failing to generalize in a systematic and robust
manner \cite{sysb-sinha-etal-2019-clutrr,sys-DBLP:journals/corr/abs-1911-02969,sysb-linzen-2020-accelerate}, (3) not learning causal and motivational comprehension \cite{definerc-dunietz-etal-2020-test}. Therefore, despite reaching great performance on many large-scale benchmarks, there remain some gaps in real-world applications.

Many efforts have been made to collect datasets to tackle above challenges, such as natural language inference \cite{glue-wang2018glue,swag-zellers-etal-2018-swag}, reading comprehension \cite{squad-rajpurkar-etal-2016-squad,race-lai-etal-2017-race}, or movie question answering using plot synopses \cite{MovieQA}. Nonetheless, natural language inference datasets \cite{glue-wang2018glue,swag-zellers-etal-2018-swag} assume sentence-level inputs, which might due to annotation expense, are not sophisticated enough for what real-world applications have to assimilate. Reading comprehension datasets \cite{squad-rajpurkar-etal-2016-squad,race-lai-etal-2017-race} lack causal or motivational queries, as pointed out by recent research \cite{definerc-dunietz-etal-2020-test}; Movie plot synopses question answering dataset, MovieQA \cite{MovieQA}, involves some casual or motivational questions (e.g. ``why” questions). However, a recent paper \cite{movieqaright-bjasani2019movieqa} suggested that MovieQA questions could be answered without reading the movie content, indicating that a learning model could reach a high score by capturing question-answer bias.  


As an epitome of the world and society, movie story comprises implicit world knowledge and various characters and their actions, motivations, and causalities
, as mentioned by recent studies \cite{MovieQA,definerc-dunietz-etal-2020-test}. Cognitive science research \cite{cogmc-johannes2018constructing} also suggests that, comparing to expository text, which is utilized for building various machine comprehension datasets \cite{edata-choi-etal-2018-quac,edata-johannes2018constructing,edata-yang-etal-2018-hotpotqa,squad-rajpurkar-etal-2016-squad}, narrative text has a close correspondence to everyday experiences in contextually specific situations. An existing dataset \cite{kar-etal-2018-mpst} uses movie synopses to predict tags associated with movies. However, most occurred tags, such as murder, violence, flashback, comedy, and romantic, could be determined with several keywords or specific patterns. Also, these tags do not involve casual and motivational comprehension.

We seek an alternative approach: \textit{trope}, which is a storytelling device, or a shortcut, frequently used in creative productions such as novels, TV series and movies to describe situations that storytellers can reasonably assume the audience will recognize. Beyond actions, events, and activities, they are the tools that the art creators use to express ideas to the audience without needing to spell out all the details. For example, \textit{``Heroic Sacrifice”} is a common trope, defined as when a character saves others from harm and is killed as a result. A film can contain hundreds of tropes, orchestrated intentionally by the production team. Some examples are shown in Figure \ref{fig:tropedet_task}. The end goal of our trope detection task is to push forward the development of a machine's cognitive skills to approach a level comparable to that of an educated human being by developing learning models and algorithms to tackle a few challenges altogether of the trope detection task.



We are optimistic that a learning system with trope detection level deep cognitive skills 
could better tackle real world tasks using long and diverse text content, including but not limited to deep comprehension, recommendation systems, chat-bot, opinion mining or personal assistant. For instance, a recommendation system with this approach could retrieve content not only with relative content but furthermore, with similar tropes (situation), which can definitely arouse users' interest. Therefore, we collect \ourdatasetfull{} (\ourdataset{})\footnote{https://github.com/htsucml/TiMoS} from a Wikipedia-style website, TVTropes\footnote{http://tvtropes.org/} with 5623 movie synopses associated with 95 most occurred tropes. The movies are diverse in genre, filming year, length, and style, making the task challenging and unable to rely on patterns from a specific domain. The tropes involve character trait, role interaction, situation, and storyline, which could be sensed by a non-expert human, but remains challenging for machines that have more than 100 million parameters and pre-trained with 11,000 books and the whole Wikipedia (23.97 F1 score while a human could reach 64.87). To take a first step towards the challenging trope detection task, we propose a new multi-stream comprehension network (MulCom), which utilizes multi-level consciousness, including word, sentence, and role relation level. We propose a novel component, Multi-Step Recurrent Relation Network (MSRRN), leveraging Recurrent Relational  Network (RRN) \cite{DBLP:conf/nips/PalmPW18} by integrating each step of representation of multi-step reasoning to utilize role interaction cues progressively. 

Our experimental results suggest that there is a considerable gap between modern learning models and human performance. Learning models, including contextual embedding \cite{devlin-etal-2019-bert}, tag prediction system \cite{kar-etal-2018-folksonomication}, and relational network \cite{DBLP:conf/nips/PalmPW18}, achieve less than 40\% human performance in terms of F1 score, indicating that trope detection is a direction worth investigating. Our MulCom outperforms all modern baselines 
by utilizing multi-level consciousness and progressively reasoning. We conduct a comprehensive analysis for ablation study, trope difficulty, knowledge adaption, and case studies. We also examine human performance, investigate human cognitive processes, and explore potential directions of our trope detection task.

To summarize, in this work, (i) we introduce a novel task -- trope detection in films, of machine comprehension. (ii) We present a new dataset \ourdataset{} (\ourdatasetfull{}) that includes 5000+ movie synopses along with their trope annotations. (iii) We provide a benchmark for this task by proposing a Multi-level Comprehension Network (MulCom) for trope detection and incorporate a GNN based network -- Multi-Step Recurrent Relational Networks (MSRNN). (iv) We comprehensively analyze the model performances and ensure the proposed dataset's answerability by leveraging a human evaluation. We also provide future research directions according to the experiment results.


\section{Related Works}

\subsection{Contextual Embedding}
Different from conventional word embedding, which assigns a fixed vector per token, contextual embedding such as ELMo \cite{elmo-Peters:2018}, BERT \cite{devlin-etal-2019-bert} and XLNet \cite{xlnet-NIPS2019_8812} captures the semantic of each word according to the context. By using self-supervised pre-training on large corpora, contextual embedding boosts learning system performance on various tasks such as machine translation, reading comprehension, text classification, language inference. For example, BERT demonstrates significant performance boost on language inference \cite{glue-wang2018glue,swag-zellers-etal-2018-swag} and reading comprehension \cite{squad-rajpurkar-etal-2016-squad,squad2-rajpurkar-etal-2018-know}. Despite the success, some limitations are excavated by recent research. \citet{bertnot-ettinger-2020-bert} discovered that BERT is insensitive to common sense, pragmatic inferences, and role-based events, which are commonplace in daily life, could easily be processed by humans, and are important for various tasks. \citet{sysb-sinha-etal-2019-clutrr} pointed out that contextual embedding models tend to incorporate the statistics rather than reasoning, and do not exhibit the systematic generalization capability. \citet{definerc-dunietz-etal-2020-test} proposed a systematic approach, Template of Understanding (ToU), to define what a learning system should comprehend. They also suggested that a competitive contextual embedding model, XLNet \cite{xlnet-NIPS2019_8812}, training on a relatively systematic dataset, RACE \cite{race-lai-etal-2017-race}, performs poor in terms of causal and motivational comprehension\footnote{ToU proposed by  \citet{definerc-dunietz-etal-2020-test} including spatial, temporal, causal and motivational comprehension, this work mainly focuses on the latter two.}. These drawbacks are gaps of applying modern learning algorithms to real world scenarios. We believe that our trope detection task could bridge this gap by examining and developing learning system in terms of consciousness, systematic generalization, and causal and motivational comprehension.

\subsection{Machine Comprehension Dataset}
Teaching machine to comprehend text content is one of the ultimate goals in natural language processing. Reading comprehension, which answers a language query according to a text passage is a widely used testbed. However, RACE \cite{race-lai-etal-2017-race}, arguably the most systematic one, as it was originally designed to examine humans, still failed to provide causal and motivational cues, as pointed out by a recent study \cite{definerc-dunietz-etal-2020-test}. Language inference is another way to evaluate comprehension capability. \citet{glue-wang2018glue} proposed a GLUE benchmark involving multiple tasks to understand beyond the detection of superficial correspondences. \citet{swag-zellers-etal-2018-swag} provided a SWAG dataset to challenge learning systems with grounded situations. While these datasets incorporate commonsense, causal and motivational inferences, they assume sentence-level inputs, which are often too simplified for real-world applications. Alternatively, our trope detection task involves conceptual and information-rich movie synopses content, together with diverse, causal and motivational tropes. Therefore, we are optimistic that trope detection task could encourage and evaluate the progress of human-level reasoning capability.


\subsection{Films Related Works}
Movie story is a proper testbed as suggested by recent machine comprehension studies \cite{MovieQA,definerc-dunietz-etal-2020-test} and cognitive science research \cite{cogmc-johannes2018constructing}. A few text-driven film analysis tasks have been presented in recent years. \citet{kar-etal-2018-mpst} presented a new task and dataset for discovering movie tags from film synopses and \citet{kar-etal-2018-folksonomication} further tackle the problem using Convolutional Neural Network (CNN) with Emotion Flow. Despite useful for movie recommendation applications, tags are relative shallow semantics that could be determined by few keywords. \citet{MovieQA} proposed a movie question answering dataset, MovieQA, including plot synopses comprehension. However, \citet{movieqaright-bjasani2019movieqa} revealed that MovieQA could be answered with question-answer pairs alone, indicating that model performance could depend on capturing language bias.

\citet{HarnessingAI} took an initiative to introduce tropes to the multimedia community, identified open research problems in this domain, and demonstrated how they will advance research in content analysis, contextual understanding, and even computer vision. Inspired by their work, we actualize this idea and formulate a new situation understanding problem.

\section{New Dataset: \ourdataset{}}\label{section:dataset}

\begin{figure*}
  \includegraphics[width=\textwidth]{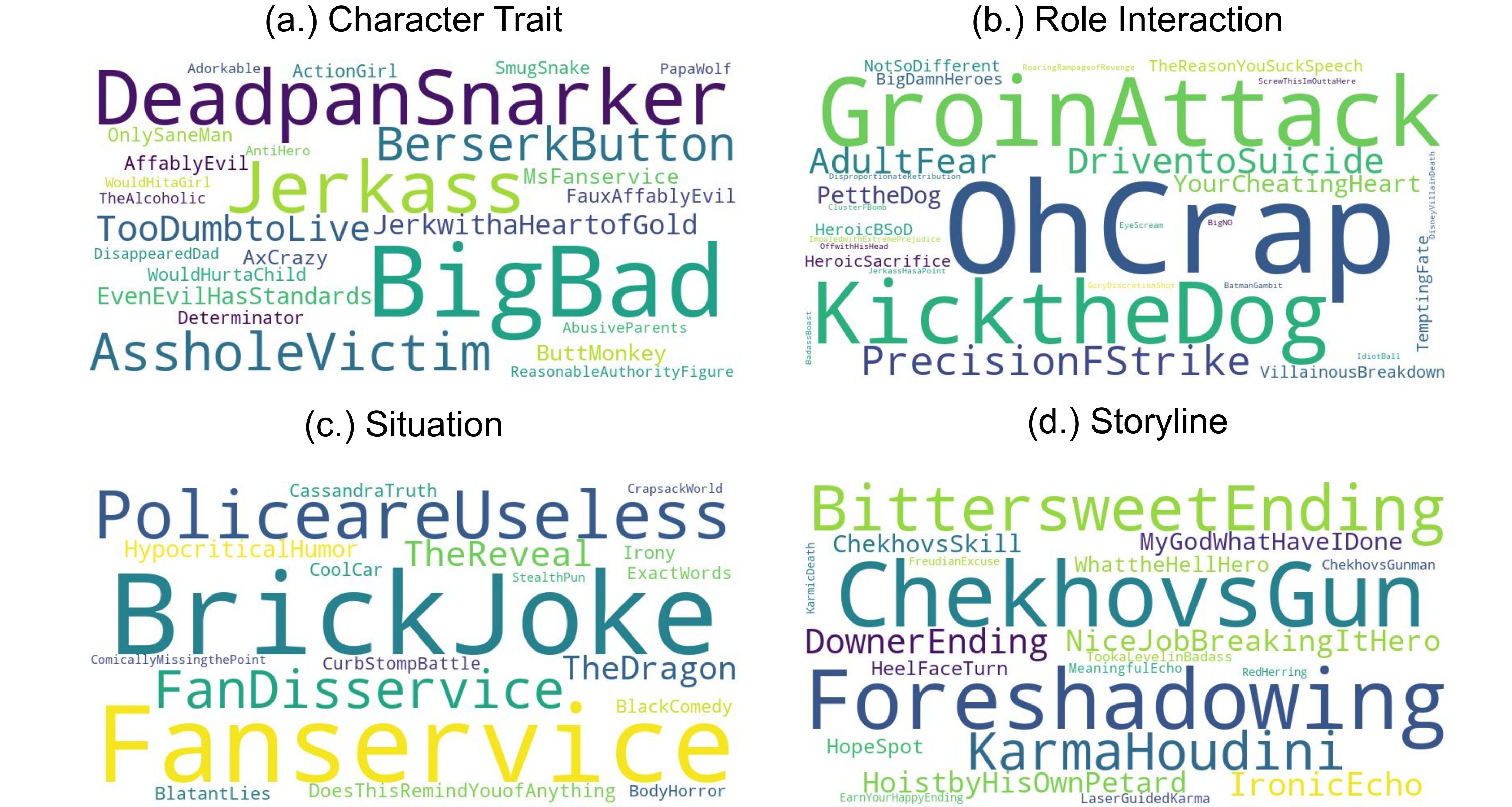}
  \caption{Word cloud of trope occurrences per category, size of the tropes in proportion to their frequency in the dataset. While the distribution of trope occurrences is skewed, the fewest trope still occurred in about 3.5\% of movies. (See Table \ref{tab:trope_statistic} for the statistics and Section \ref{section:dataset} for detailed discussion.)}
  \label{fig:tropecloud}
\end{figure*}
\subsection{Overview}\label{subsec:dataset:overview}
We present a novel dataset \ourdataset{} (\ourdatasetfull{}) which includes (1) 5623 movie synopses, 
(2) 95 tropes 
along with their definitions and human labeled categories (3) trope appearances in each film. Since the tropes are very diverse, we classify them into 4 categories. Figure \ref{fig:tropecloud} shows a trope cloud for each category, with size of the tropes in proportion to their frequency in the dataset. Table \ref{tab:movie_statistic} and \ref{tab:trope_statistic} shows statistics about the dataset.

\textbf{Character Trait} tropes describe personalities, strengths, or weaknesses. A character trait may be more complicated than ``good” or ``evil,” such as \textit{Even Evil Has Standard}, which keeps villains safely on the ``still sympathetic” side, but still has some distance to be good.

\textbf{Role Interaction} tropes depict role awareness, encounter, or actions in the story. A role interaction might be associated with character traits of the role, e.g., \textit{Pet the Dog} is when villains engage in a moment of kindness to demonstrate that they are not that bad after all.

\textbf{Situation} tropes describe scene level scenarios where some concepts or emotions are delivered to audiences using certain entities, actions, or role interactions. A situation could be connected to a concrete object such as \textit{Cool Car}, or an abstract one such as \textit{Black Comedy}.

\textbf{Storyline} tropes control directions of the movie by turning, foreshadowing, echoing or ending  the story by a scene (\textit{Hope Spot}), an object (\textit{Chekhov's Gun}) or simply an attempt (\textit{Red Herring}).


\subsection{Data Collection}
\paragraph{Movie Synopses}
We obtain movie synopses from the MPST data\-set \cite{kar-etal-2018-mpst}, where the synopses are collected from Internet Movie Database (IMDb) and Wikipedia. Synopses are crawled from Wikipedia when plot synopses are unavailable from IMDb, or the synopses in Wikipedia were longer than the synopses in IMDb. 
Each synopsis is written chronologically to preserve the temporal relation between events, which is crucial as many tropes could only be revealed with multiple events and their temporal order in the movie. Every synopsis has at least 8 \footnote{Original MPST dataset synopsis has at least 10 sentences. We further clean the dataset and remove some sentences with only a quote mark.} sentences and provides adequate details about every important character and his/her/its intention, emotion, and interaction with other roles. 

\paragraph{Trope Collection}
We collect tropes and their definitions on a Wikipedia-style database, TVTropes, where the definition, examples, and occurrences in movies are annotated by web users, so-called tropers. Specifically, we query TVTropes with the movie titles from MPST movie list to obtain the associated tropes of the corresponding movie. 

\paragraph{Trope Selection}
First, we include 144 tropes that appeared the most in plot synopses as models require adequate data to learn. Next, we manually select 95 tropes which are the most relevant to the story, as some tropes do not focus on the movie itself (e.g., \textit{Creator Cameo} focuses on the creator, the director or the producer; \textit{Soundtrack Dissonance} is defined by the music) or requires external knowledge source (e.g., the cover for \textit{Covers Always Lie}), making them impossible to detect with movie synopses. Finally, we split the dataset of 5623 synopses into train (4059/5623, 72.19\%), val (715/5623, 12.72\%), and test (849/5623, 15.09\%) set. We briefly split 49 unused tropes into 6 categories, including references to previous works, adaptation approach, storytelling/shooting techniques, producers' abilities, reference to real-word knowledge, and narrative conventions, and leave them to future research such as multi-document trope detection.


\subsection{Data Analysis}\label{section:dataset:analysis}


\paragraph{Statistics} Table \ref{tab:movie_statistic} summarizes movie synopses statistics, the distribution of trope occurrence, sentence count, word count, roles, and co-references (corefs) are skewed. We use Spacy \cite{spacy2} for sentence segmentation and word tokenization, and a co-reference resolution tool \cite{clark-manning-2016-deep, clark-manning-2016-improving} to get roles and corefs.
While few synopses are really large (with more than 1,000 sentences) or contain many tropes (cover more than 70\% of tropes), half of the synopses have less than 40 sentences or less than 10 tropes. Trope occurrence distribution (Table \ref{tab:trope_statistic}) is also skewed, but with at least 200 (4\%) appearance, allowing learning models to capture the meaningful context. 

\begin{table}[ht]
\centering
\noindent
\begin{tabular}{llccccc}
\toprule
&&Median & Average & Min & Max & \(\sigma\) \\
\midrule
\multirow{5}*{Train}
& Tropes & 8 & 11 & 1 & 68 & 10   \\
& Words & 926 & 1326 & 141 & 12436 & 1229   \\
& Sentences & 39 & 59 & 8 & 666 & 61   \\
& Roles &25 & 33 & 0 & 273 & 28    \\
& Corefs  &131  &197  &0 &2307 &208  \\
\midrule
\multirow{5}*{Val}
& Tropes & 8 & 12 & 1 &76 & 10  \\
& Words  & 913 & 1294 & 175 & 15099 & 1259  \\
& Sentences & 37 & 57 & 8 & 1181 & 69  \\
& Roles & 24 & 32 & 2 & 290 & 28    \\
& Corefs  &126  &189  &15 &1959 &203  \\
\midrule
\multirow{5}*{Test}
& Tropes  & 9 & 12 & 1&69 & 10   \\
& Words & 921 & 1305 & 180 & 11712 & 1163  \\
& Sentences & 39 & 58 & 9 & 757 & 60   \\
& Roles & 25 & 32  &3  & 231 & 26 \\
& Corefs &127  &194  &13  &1881 &198 \\
\bottomrule
\end{tabular}

\caption{\ourdataset{} dataset movie synopsis statistics. It shows the statistic of sentence/word length and number of tropes' occurrence. The average synopsis length is long and standard deviation is quite large.  (\(\sigma\) denotes standard deviation.) (cf. Section \ref{section:dataset:analysis})}
\label{tab:movie_statistic}
\end{table}

\begin{table}[ht]
\centering
\noindent

\begin{tabular}{lccc}
\toprule
\multicolumn{4}{c}{\% occurrence in movie synopses}\\
\midrule
        & Train  & Val   & Test           \\
\midrule 
Median  & 6.60\% & 6.64\%   & 7.03\%        \\
Average & 7.84\% & 8.04\% & 8.00\%     \\
Minimum & 4.34\%  & 3.50\%   & 3.63\%        \\
Maximum & 32.10\%   & 33.85\% & 34.06\%      \\
\(\sigma\)  & 3.96\%     & 4.21\%     & 4.10\%          \\
\bottomrule
\end{tabular}
\caption{The statistic of tropes' occurrence in movies in percentage. While the distribution is skewed, the fewest trope occurs in 3.5\% of movies, and enables learning systems to capture meaningful context.(cf. Section \ref{section:dataset:analysis})}
\label{tab:trope_statistic}
\end{table}


\paragraph{Correlation} 
Table \ref{tab:cooccu} demonstrates top co-occurrences calculated using the Intersection over Union (IoU) metric, i.e., the ratio of trope co-occurrence and trope occurrence, respectively. The list shows that tropes related to violent scenes such as \textit{Impaled with Extreme Prejudice}, \textit{Off with His Head!}, and \textit{Eye Scream} are often used in the same movie. Also, the co-occurrence of \textit{Hypocritical Humor}, \textit{Comically Missing the Point}, and \textit{Stealth Pun} are frequently linked together to make a joke. Finally, \textit{Would Hurt a Child} might be the cause of \textit{Papa Wolf}. 

\subsection{Challenges}\label{sec:dataset:challenges}
We further analyze the key attributes of tropes, how they bring challenges to the task, and how trope detection capability enhances current learning systems.

\paragraph{Consciousness}, which digests and distills information from long and information-rich content and capture essential elements to complete the task, is crucial for various learning systems. Recently, consciousness is mostly formulated as several attention mechanisms, such as context-query attention \cite{bidaf} or self-attention \cite{attentionisallyouneed,devlin-etal-2019-bert}, and achieve great successes in many tasks such as document classification, machine translation and reading comprehension. Most of these approaches focus on short, single domain documents and a single task. A recent paper \cite{bertnot-ettinger-2020-bert} suggests that BERT \cite{devlin-etal-2019-bert} is less sensitive to role-based events, which are common is daily life and real-world data. Additionally, real-world content such as blogs, forum posts, or social media, are naturally long, multi-domain, and not task-specific. Therefore, tropes, which focus on role interactions, diverse in their contents, and expressed in multiple ways (objects, actions, emotions, or interactions), are closer to real-world scenarios.

\paragraph{Systematic Generalization}, which refers to known components, such as tokens in language processing, and produce an infinite number of new combinations. Systematic Generalization is important to handle data in practice where (1) Human written or spoken data could be conveyed in almost infinite combinations. (2) Language is ever-changing, especially in the web age, where netizens create and combine new words worldwide. More than 550 new words are added into the Oxford English Dictionary in March 2020 bimonthly update\footnote{https://public.oed.com/updates/}. Many of these new words are new compositions of words or sub-word units, making humans sense the meaning with few observations and little context. However, neural network models, despite being powerful on various benchmarks, still show little systematic generalization capability, as pointed out by recent works \cite{sys-bahdanau2018systematic,sys-DBLP:journals/corr/abs-1911-02969,sys-ICML-2018-LakeB,sysb-goodwin-etal-2020-probing,sysb-sinha-etal-2019-clutrr}, even with pre-trained contextual embedding such as BERT \cite{sysb-sinha-etal-2019-clutrr,sys-DBLP:journals/corr/abs-1911-02969}. Various tropes manipulate words and their combinations, sequences, co-occurrences, and analogies to trigger audiences' emotions, such as a knowing smile, which are good for testing systematic generalization ability. 

\paragraph{Causal and Motivational Comprehension}, which refers to the root cause of certain events or state changes in the story \cite{definerc-dunietz-etal-2020-test}, is necessary for investigating the evidence in data mining tasks. To determine if some events or intentions lead to the corollary, or simply a coincide. While modern machine comprehension models \cite{qanet-wei2018fast,devlin-etal-2019-bert,xlnet-NIPS2019_8812} performs human-level comprehension ability on various large-scale, human-annotated benchmarks such as SQuAD \cite{squad-rajpurkar-etal-2016-squad} or RACE \cite{race-lai-etal-2017-race}, a recent research \cite{definerc-dunietz-etal-2020-test} suggests that these leaderboard-dominating models are not learning crucial causal or motivational elements. Tropes incorporate many casual and motivational elements behind actions and render them meaningful, making them a good test-bed for testing comprehension.

\begin{table}
\centering
\begin{tabular}{cc}
\midrule
Disney Villain Death & The Dragon \\ \hline
Comically Missing the Point & Hypocritical Humor \\ \hline
Impaled with Extreme Prejudice & Off with His Head! \\ \hline
The Reveal & Red Herring \\ \hline
Hypocritical Humor & Stealth Pun \\ \hline
Badass Boast & Curb-Stomp Battle \\ \hline
Eye Scream & Off with His Head! \\ \hline
Would Hurt a Child & Papa Wolf \\
\midrule
\end{tabular}
\caption{Top trope co-occurrence pairs (two trope co-occured in the same movie) in \ourdataset{} dataset. It shows that some tropes are used together to deliver some story. (cf. Section \ref{section:dataset:analysis})}
\label{tab:cooccu}
\end{table}


\section{Method}\label{section:method}
To spearhead the development of the trope detection task, we tackle the challenges mentioned in Section \ref{sec:dataset:challenges} with two novel designs: Multi-Level Comprehension Network (MulCom, Figure \ref{fig:framework}, Section \ref{section:method:mulcom}), and Multi-Step Recurrent Relational Network (MSRRN, Figure \ref{fig:gnn}, Section \ref{section:method:msrrn})

\subsection{Task Formulation}
We formulate the task as a multi-label classification problem. Given a film, the model is asked to predict the tropes' appearance in the film. More specifically, the input instance is a film synopsis, and the output is a binary vector with size \(\lvert T \rvert\), where \(T = \{t_1, t_2, ...\}\) denotes the set of all the tropes. 

\subsection{Multi-Level Comprehension Network}\label{section:method:mulcom}
\begin{figure*}
 \center
  \includegraphics[width=\textwidth]{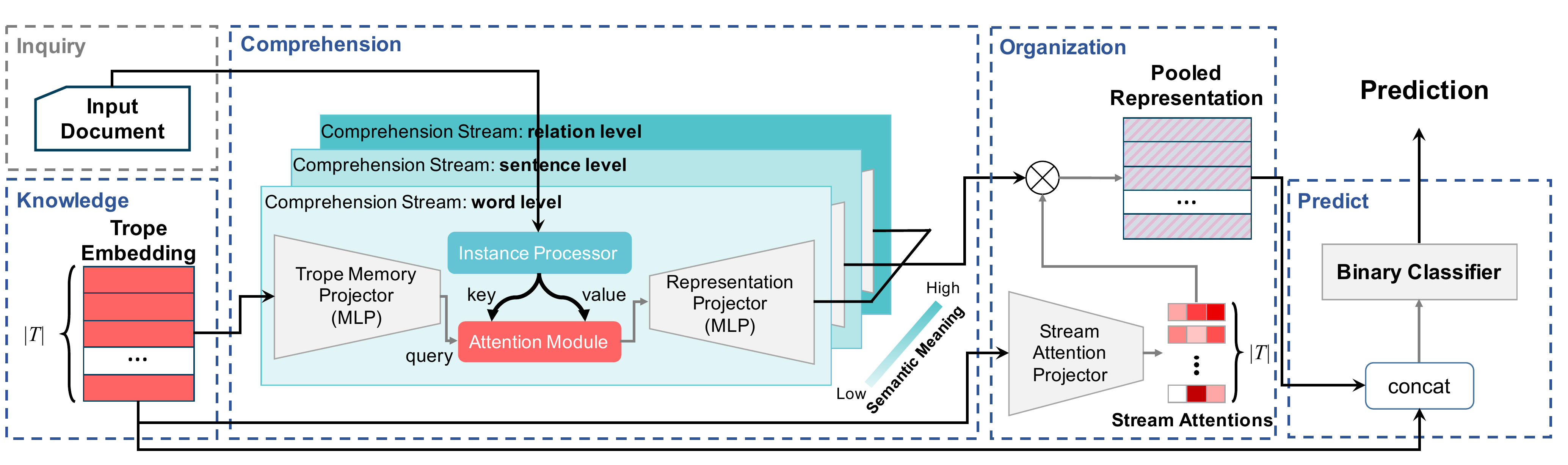}
  \caption{Multi-Level Comprehension Network (MulCom). 
  The MulCom is designed to deal with different levels of cues in movie synopses with multiple streams. (1) \textit{Inquiry} processes the input movie synopsis document and extracts the features. (2) \textit{Knowledge} embeds each trope. (3) \textit{Comprehension} leverages multiple levels of signals, from low semantic words to high semantic role relations. First, the module projects trope embedding and movie synopsis to the same space with Trope Memory Projector and Instance Processor, respectively. Then, it focuses on movie synopsis according to tropes by attention mechanism. Finally, the fused representation is projected by another MLP. In this work, we design a novel component Multi-Step Recurrent Relational Network (MSRRN, Figure \ref{fig:gnn}, Section \ref{section:method:msrrn}) for relation level stream.
  4. \textit{Organization} integrates signals from different levels according to stream attentions, which is projected from Trope Embedding. (5) \textit{Predict} concatenates multi-level movie synopsis and each trope embedding and predict if a trope occurs.
  (cf. Section \ref{section:method:mulcom})
  }
  \label{fig:framework}
\end{figure*}

Seeing the content diversity and abundance in tropes of different categories, we believe each trope is best detected by paying attention on different level of cues.
For example, trope occurrence is usually sensitive to important lexical cues, such as \textit{Pet the Dog} might occurs when a character is doing violence. However, the motivation which makes the violence cruel might require role or role interaction level signals, such as the behavior of the victim or certain interactions. 
Therefore, we designed a multi-stream network where each stream performs a different level of comprehension, and each trope is detected using a different combination of proportions among the streams. The architecture of our Multi-Level Comprehension Network (MulCom) is shown in Figure \ref{fig:framework}. The model can be divided into three parts: \textit{Knowledge}, \textit{Comprehension} and \textit{Organization}.

\paragraph{Knowledge.}
The network contains a \textit{Trope Embedding} matrix \(\boldsymbol{E}\in\mathbb{R}^{\lvert T \rvert\times d_f}\), where \(T\) is the trope set, that serves as the memory of the model and stores its knowledge about tropes. Each row of \(\boldsymbol{E}\) represents a trope's features which will be used as query to perform attention mechanism on the incoming input document in the following \textit{Comprehension} step. \(\boldsymbol{E}\) is randomly initialized in the beginning of training and will be updated throughout training to ultimately acquire a meaningful representation for each trope. The existence of Trope Embedding allows the model to perform attention using trope as queries without any prior knowledge about tropes. 

\paragraph{Comprehension.}
In this step, the network performs different levels of comprehension skills on the input document and uses its knowledge about tropes from Trope Embedding to focus on relevant content. In order to perform multiple levels of comprehension, the network contains numerous \textit{Comprehension Streams} with each performing a different level of comprehension. More specifically, each stream has a unique component called \textit{Instance Processor} that processes the input document and obtains a sequence of document features. We use Word2Vec \cite{mikolov2013distributed} word embedding, BERT \cite{devlin-etal-2019-bert} sentence encoder and a GNN based Multi-Step Recurrent Relational Network as the Instance Processor of each stream to perform \textit{word level, sentence level and relation level} of comprehension, respectively. Finally, Trope Embedding is used as a query to perform attention mechanism on document features in \textit{Attention Module} to obtain a relevant representation \(r_{t}^s \) for each trope where \(s\) denotes the Comprehension Stream \(s\) and \(t\) denotes the trope \(t\).

\paragraph{Organization.}
In this step, the network organizes representations from different levels of Comprehension Streams by using a MLP called \textit{Stream Attention Projector} and feeds \(\boldsymbol{E}\) as input to obtain stream attentions \(\boldsymbol{A}\). This can be viewed as using the network's knowledge about tropes to determine the importance of each Comprehension Stream with respect to each trope. The network then aggregates the representations \(R = \{r_t^{word}, r_t^{sentence} , r_t^{relation}\}\) for each trope \(t\) using stream attention:
\[\boldsymbol{A} = StreamAttentionProjector(\boldsymbol{E}), \;\;\;\; r_{t} = \sum_{s \in S} \boldsymbol{A}_{t,s} \cdot r_t^s,\]
where \(A \in\mathbb{R}^{\lvert T \rvert\times\lvert S \rvert}\) is the attention matrix and \(r_t\) is the representation for trope \(t\) that integrate different level of comprehension. 

\paragraph{Predict.}
The network links \(r_t\) with its knowledge of tropes from Trope Embedding by concatenating them, and sends them into a binary classifier to predict if the trope appear in the input document.

\paragraph{Loss Function.}
We adopt \textit{Binary Cross Entropy} as our loss function:
\[L = \frac{1}{B} \sum_{b=1}^{B} \frac{1}{T} \sum_{t=1}^{T} -[p \cdot y_{b,l} \cdot \log (\sigma(x_{b,t})) + (1-y_{b,t}) \cdot \log (1-\sigma(x_{b,t}))],\]
where \(B\) is the batch size, \(p\) is the weight for positive prediction, \(\sigma\) represents Sigmoid function, and \(y_{b,t}\) and \(x_{b,t}\) is the ground truth and prediction of trope \(t\) and \(b\)th data in the batch.

\subsection{Multi-Step Recurrent Relational Network}\label{section:method:msrrn}

\citet{bertnot-ettinger-2020-bert} highlighted that one of the limitations of BERT model is being insensitive to role-based events. \citet{sysb-sinha-etal-2019-clutrr} also exhibited stronger generalization and robustness of graph-based model comparing to contextual embedding such as BERT. 
To process trope signals, apart from token and sentence level ones which are explored by many recent studies, we resort to a GNN based model here because we believe that receiving in-depth knowledge from a film story relies heavily on the understanding of characters' interaction, and GNNs is able to preserve and reason through the interaction of entities.

\begin{figure}
  \includegraphics[width=\columnwidth]{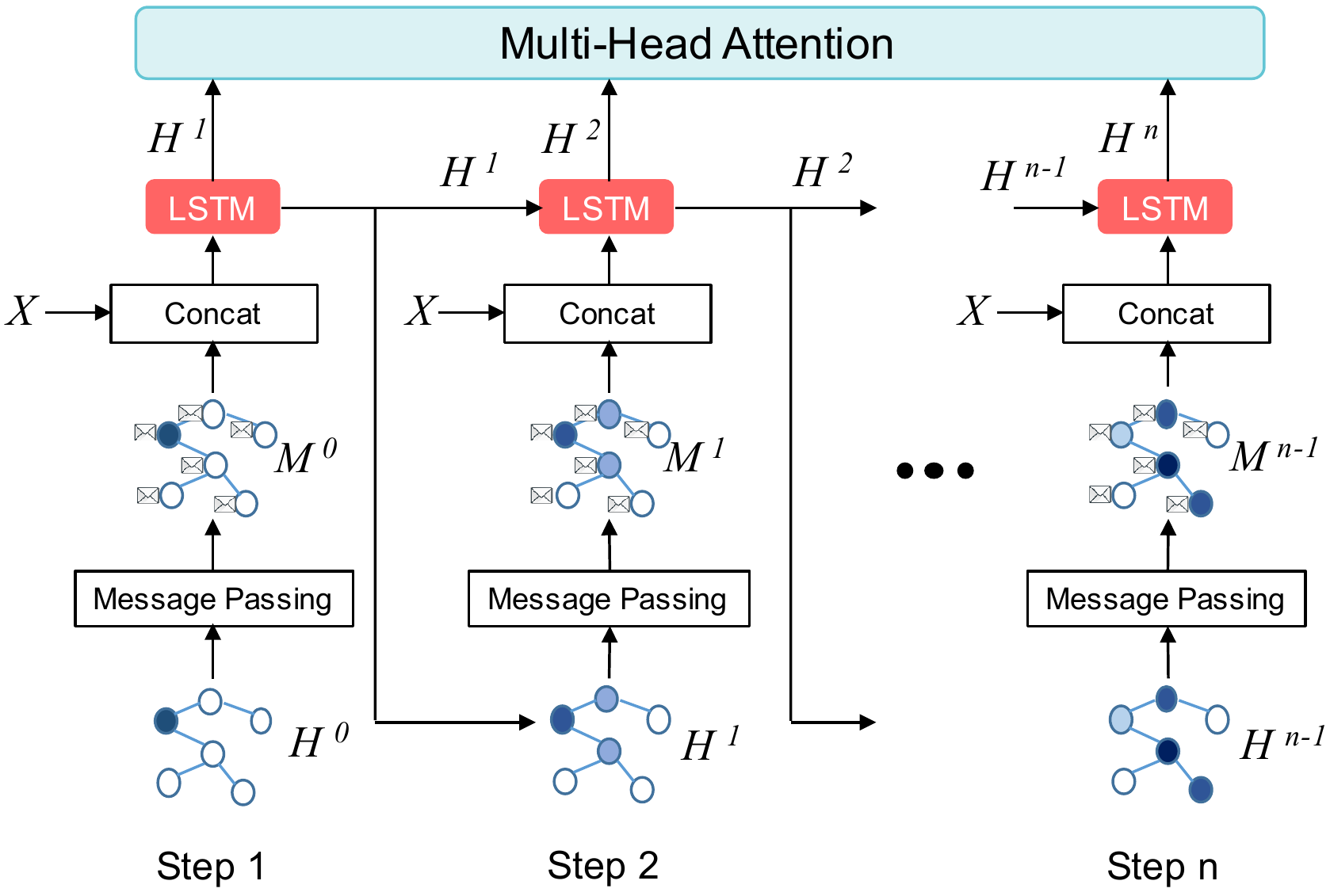}
  \caption{Multi-Step Recurrent Relational Network (MSRRN, relational level instance processor in Figure \ref{fig:framework}: Comprehension). Different from RRN \cite{DBLP:conf/nips/PalmPW18}, which only uses the \textit{last step} representation, MSRRN performs multiple steps of reasoning and fuses the hidden states from \textit{all steps}. Each reasoning step takes in a graph with nodes hidden values \(H\) and performs Message Passing to obtain messages \(M_{step}\) to each node from its neighboring nodes. The messages are then concatenated with the node features and sent into a LSTM as its input to obtain a new hidden value \(H_{step+1}\) for each node. The hidden values \(H_{step}\) of each reasoning step are merged using Multi-Head Attention. (cf. Section \ref{section:method:msrrn})}
  \label{fig:gnn}
\end{figure}

Utilizing an existing GNN model such as RRN might be a straightforward way. However, RRN \cite{DBLP:conf/nips/PalmPW18} applies multi-step reasoning but only utilizes the output representation of \textit{the last step}. Therefore, to comprehend the causality and the motivation in a movie synopsis,  we present a Multi-Step Recurrent Relational Network (MSRRN), which tracks \textit{each reasoning step} by applying self-attention on hidden values from all steps to preserve the vital information. The architecture is shown in Figure \ref{fig:gnn}.

\paragraph{Preprocess.}
In order to convert a film synopsis into a graph, we use coreference resolutions tools \cite{clark-manning-2016-deep, clark-manning-2016-improving} to extract entities in film synopses. These entities are then encoded by BERT \citet{devlin-etal-2019-bert} and used as node features. We use all the sentences that contain \(node_i\) and \(node_j\)
to represent the \textit{``implicit relation”} between them. To be more specific, the features of the edge connecting the two nodes is the summation of the BERT encoded sentences that contain them. A node can have self-loop if it is the only entity appearing in the sentence. The average node and graph density in the dataset is 16.2 and 0.38.

\paragraph{Reasoning}
MSRRN takes in a graph as input and iteratively updates each node using information propagated from its neighboring nodes. We call this update process \textit{Reasoning}.

Nodes \(N = \{n_1, n_2,...\}\) has node features \(X = \{x_1, x_2,...\}\) and hidden values \(H^t = \{h_1^t, h_2^t,...\}\) for each reasoning step \(t\).  
The network performs \textit{Message Passing} by calculating a \textit{message} \(m_i^t\) for \(n_i\) that encompasses the information from connected nodes (neighbors) and their relations (edge feature): 
\[m_{ij}^t = MLP(e_{ij}, h_i^t, h_j^t),\;\;\;\; m_i^t = \sum_{j\in N(i)}m_{ij}^t,\]
where \(m_{ij}^t\) is the message propagated from \(n_j\) to \(n_i\) in reasoning step \(t\),  \(N(i)\) denotes the neighboring nodes and \(m_i\) is the aggregated messages propagated from all the neighboring nodes. The network then updates \(h^t\) using Long Short-Term Memory (LSTM) \cite{hochreiter1997long} and obtain a new hidden value \(h_i^{t+1}\).
\[h_i^{t+1} = LSTM(h_i^t,concat(x_i, m_i^t)),\]
where \(h_i^t\) and \(concat(x_i, m_i^t)\) serve as the hidden state and input for the LSTM, respectively.

\paragraph{Self-Attention on Multiple Reasoning Step}
After the reasoning process, we obtain a hidden value sequence \(\Omega =(H^1,H^2,..., H^T)\) where \(T\) is the number of total reasoning step. We use Multi-Head Attention proposed by \citet{NIPS2017_7181} to apply self-attention on \(\Omega\). 

Here, since we are performing self-attention, The query, key and value are all \(\Omega\). We sum up all the element in self-attended sequence \(\Omega^\prime\) and obtain the final hidden value sequence \(\tilde{H}\):
\[\Omega^\prime = MultiHead(\Omega,\Omega,\Omega),\;\;\;\tilde{H} = \sum_{i=1}^T \tilde{H}_i,\; \tilde{H}_i\in\Omega^\prime.\]

\section{Experiments}\label{section:experiments}
We conduct experiments on the dataset \ourdataset{}. We examine the performance of our proposed Multi-Level Comprehension Network (MulCom) and three competitive baseline systems.\footnote{https://github.com/htsucml/TiMoS} 
We evaluate the models using F1 score (F1) and mean average precision (mAP). These metrics take both recall and precision into consideration.

\subsection{Baselines}\label{section:experiments:baselines}
Our experiment uses the following models:

BERT \cite{devlin-etal-2019-bert} and its variants have been state-of-the-art for many NLP tasks. We use BERT-base with 12 layers in our experiments. As movie synopses are usually much longer than 512 tokens, cutting off the rest of the text would lead to information loss. We fine-tune BERT on our dataset by segmenting long film synopses into paragraphs, each containing less than or equal to 128 words. All encoded paragraphs are then pooled and sent into a binary classifier so that the BERT model can capture the full context of movie synopses.
  
Folksonomication \cite{kar-etal-2018-folksonomication} is a modern movie tags prediction model that initially runs a multi-tag classification task on movie synopses. The model predicts a confidence score for each label. However, instead of setting a threshold for confidence score, they set a threshold for labels' ranking. They assume all the films have the same amount of positive labels (which is not true) and select the labels with top\(x\) confidence score as positive predictions. We follow their configuration and set \(x\) as the average label in the training dataset when evaluating. We also modify the evaluation approach by setting a reasonable threshold on the confidence score.

Recurrent Relational Network (RRN) \cite{DBLP:conf/nips/PalmPW18} is a modern graph-based model that captures relation context with recurrent neural networks. The main difference between RRN and our MSRRN is that RRN only uses the last step representation, while our model integrates all reasoning steps.

\subsection{Streams}
We experiment with different comprehension streams on our Multi-Level Comprehension Network (MulCom, Section \ref{section:method:mulcom}). The following are the streams we use in our experiment:

\textit{Multi-Step Recurrent Relational Network Stream:} The Instance Processor is Multi-Step Recurrent Relational Network proposed by us (cf. Section \ref{section:method:msrrn}).

\textit{Sentence Vector Stream:} The Instance Processor of this stream is a fixed BERT with RNN. The fixed BERT is used to extract sentence features from film synopses. The sentence feature sequences are then further encoded by a RNN. 

\textit{Word Vector Stream:} Similar to Sentence Vector Stream, except here we use Word2Vec \cite{mikolov2013distributed} to extract word features. And the word feature sequences are not further encoded by a RNN because the long sequence length might cause gradient vanishing/explosion.

\begin{table}[ht]
    \centering
   \begin{tabular}{lrr}
 \toprule  
 Baseline (Section \ref{section:experiments:baselines}) & F1 & mAP \\
 \midrule
 Random  &13.97 &8.14  \\
 BERT \cite{devlin-etal-2019-bert} (frozen)  &  23.22  & 16.24 \\
 BERT \cite{devlin-etal-2019-bert} (fine-tuned)  & 23.97  & 17.26 \\
 Folksonomication \cite{kar-etal-2018-folksonomication} & 19.93 &15.82 \\
 Folksonomication \cite{kar-etal-2018-folksonomication} top7 &20.94 &16.44 \\
 Folksonomication \cite{kar-etal-2018-folksonomication} FastText  & 22.53 & 16.35 \\
 Folksonomication \cite{kar-etal-2018-folksonomication} FastText top7 &21.27 &16.83 \\
 RRN \cite{DBLP:conf/nips/PalmPW18} & 23.20 & 15.90\\
 \midrule
 MulCom (Ours, Section \ref{section:method:mulcom}) & F1 & mAP \\
 \midrule
 MSRRN (Section \ref{section:method:msrrn}) & 23.53 & 16.90 \\
 MSRRN+Word & 23.67 & 16.31  \\
 MSRRN+Sent & 23.93 & 17.41 \\
 MSRRN+Word+Sent & \textbf{25.00} & \textbf{18.73} \\
 
 \bottomrule
 \end{tabular}
   \caption{Evaluation results on \ourdataset{}. It shows the percentage of F1 score and mAP for each method. The first block shows that even for advanced learning systems, trope detection is still challenging. 
   The second block demonstrates the performance of our proposed method. Our novel component MSRRN (Section \ref{section:method:msrrn}) surpasses RRN by utilizing the signals from all reasoning steps.
   Our full MulCom (Section \ref{section:method:mulcom}) outperforms all baselines by fusing different levels of cues, as shown in the third block. Section \ref{section:experiments} for settings and Section \ref{section:analysis} for analysis.}
   \label{tab:performance}
\end{table}

\section{Results and Analysis}\label{section:analysis}
\subsection{Baseline Comparison}
The first block of Table \ref{tab:performance} examines the trope detection performance of modern learning systems. Clearly, none of the existing models reaches significant performance, even powerful contextual embedding model BERT \cite{devlin-etal-2019-bert}. It reveals that trope detection involving deep cognitive skills is still challenging and worth exploring. Also, in comparison with frozen BERT, fine-tuned BERT is 0.75 F1 and 1.06 mAP better. A tag prediction system \cite{kar-etal-2018-folksonomication} gets at most 22.53 F1 (FastText) or 16.83 mAP (FastText top7). On the MPST movie tag prediction task, this system can reach an F1 score of about 37.0. Considering our \ourdataset{} and MPST both use IMDB movie synopses and perform multi-label classification tasks, the performance gap between tag prediction and trope detection (22.53 and 37.0, roughly 39\% drop) implies that trope detection is a more challenging task. RRN \cite{DBLP:conf/nips/PalmPW18} using relation signals instead of plain text, reaches performance comparable to frozen BERT, without 100 million scale parameters and pre-training.

The second block of Table \ref{tab:performance} compares different settings of our MulCom model (third block). Our MSRRN outperforms RRN \cite{DBLP:conf/nips/PalmPW18} in terms of both F1 and mAP by fusing all steps of representation of multi-step reasoning. Our MulCom achieves the best performance in terms of both F1 and mAP metrics by augmenting word and sentence streams, demonstrating that combining cues in different representation levels is beneficial for deep cognitive tasks such as trope detection. 

\subsection{Trope Difficulty}\label{section:analysis:diff}
Table \ref{tab:diff} demonstrates the 5 hardest and easiest tropes for models according to our experimental results. For the hardest ones, we report the maximum F1 of all models in Table \ref{tab:performance}; For the easiest ones, we report the minimum F1. According to Table \ref{tab:performance}, we see roughly 20 of the F1 score gap between the hardest and the easiest tropes, random guess level performance (13.97, according to Table \ref{tab:performance}) for certain tropes. \textit{Would Hurt a Child}, which is often used to demonstrate a character's brutality, is barely detected (12.36). Another trope describing a literally similar concept, \textit{Kick a Dog} (A role doing something evil for no gain, to show their cruelty.), has above average F1 score. Because \textit{Would Hurt a Child} is usually exhibited in a more veiled way, children abuse in films or literature is considered dark and controversial.
Additionally, this trope becomes tricky when children themselves are creepy or even evil. Some seemingly abstract tropes are easy to detect by machines, such as \textit{Chekhov's Gun} and \textit{Foreshadowing}, which might be detected by story patterns associated with these tropes with sufficient enough examples (See Figure \ref{fig:tropecloud}). \textit{Big Bad} and \textit{Deadpan Snarker} are usually main characters and can be captured with ample signals in the plot.

\begin{table}[ht]
\centering
\noindent
\begin{tabular}{llc}
\toprule
&&F1 \\
\midrule
\multirow{5}*{Easy Tropes}
& Chekhov's Gun &  38.58   \\
& Oh, Crap! & 38.50   \\
& Foreshadowing  &  37.99 \\
& Big Bad  &  36.07 \\
& Deadpan Snarker  &  33.61  \\
\midrule
\multirow{5}*{Hard Tropes}
& Would Hurt a Child  &  12.36 \\
& Blatant Lies  & 12.95 \\
& Cassandra Truth  &  13.73 \\
& Smug Snake & 14.97  \\
& Laser-Guided Karma  &  15.22  \\

\bottomrule
\end{tabular}
\caption{5 easiest and hardest tropes according to our experimental results. While learning systems could capture some trope patterns, many tropes remain challenging and result in a random guess level score (13.97) for all models. (cf. Section \ref{section:analysis:diff})}
\label{tab:diff}
\end{table}

\subsection{Knowledge Adaption}\label{section:analysis:adapt}
\begin{figure}
    \centering
    \includegraphics[width = \columnwidth]{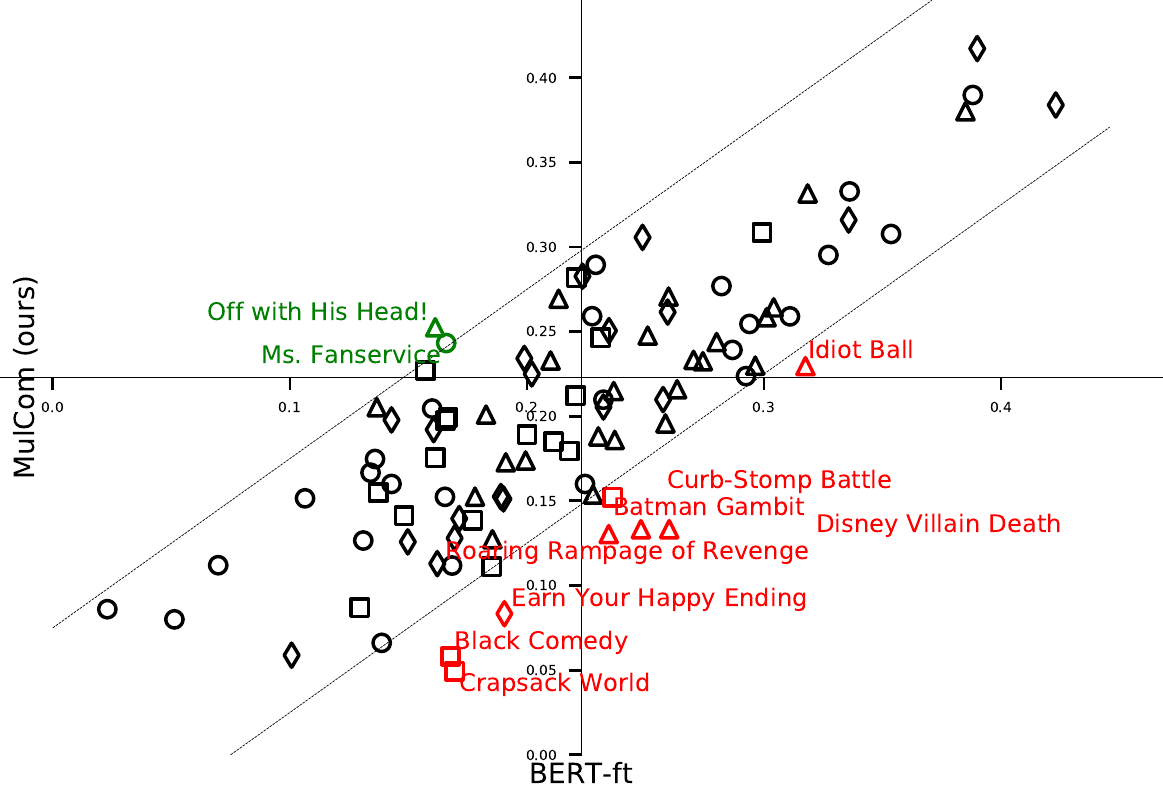}
    \caption{Per trope F1 comparison of BERT (fine-tuned) and MulCom (full). $\circ$: Character Trait, $\triangle$: Role Interaction, $\square$: Situation, $\diamond$: Storyline. X-axis: MulCom F1 score, Y-axis: BERT-ft F1 score. Red (Green): MulCom (BERT-ft) is significantly better.
    Our model outperforms BERT in various role interaction and situation tropes. The comparison also suggests the strengths and weaknesses of different knowledge adaption approaches (MulCom: Feature Extraction and BERT-ft: Fine-tuning).(cf. Section \ref{section:analysis:adapt})}
    \label{fig:mcbert}
\end{figure}

Our proposed MulCom and BERT fin-tuning baseline adopt two different ways to adapt pre-trained knowledge: \textit{Feature Extraction} and \textit{Fine-tuning}. Both ways are common and result in comparable performance on classification tasks, as pointed out by a recent research \cite{tuneornot}. Figure \ref{fig:mcbert} compares the per trope F1 between our proposed MulCom, and BERT fine-tuned (BERT-ft) baseline. We highlight tropes that MulCom (red) or BERT-ft performs significantly better (F1 gap > 7.5). We observe that (1) Our method significantly outperforms fine-tuned BERT in 8 types of tropes (bottom-right corner) while losing 2 types of tropes (top-left corner). (2) Our MulCom model performs better on role interaction ($\triangle$) and situation ($\square$) tropes. (3) Performance gaps in some categories reveal that our model tackles deep cognitive challenges. For example, 
\textit{Earn Your Happy Ending} is not only a good end but also hard work to pass an anguish journal, which renders the ending happy and worth cherishing. To determine if the protagonists \textit{earn} their happy ending , the model needs to comprehend the cause of the positive outcome.
(4) As different knowledge adaption approaches have their advantage, we see the potential to combine two approaches and point out an exciting direction for future research.

\subsection{Qualitative Analysis}\label{section:analysis:quality}
\begin{figure}
  \includegraphics[width=\columnwidth]{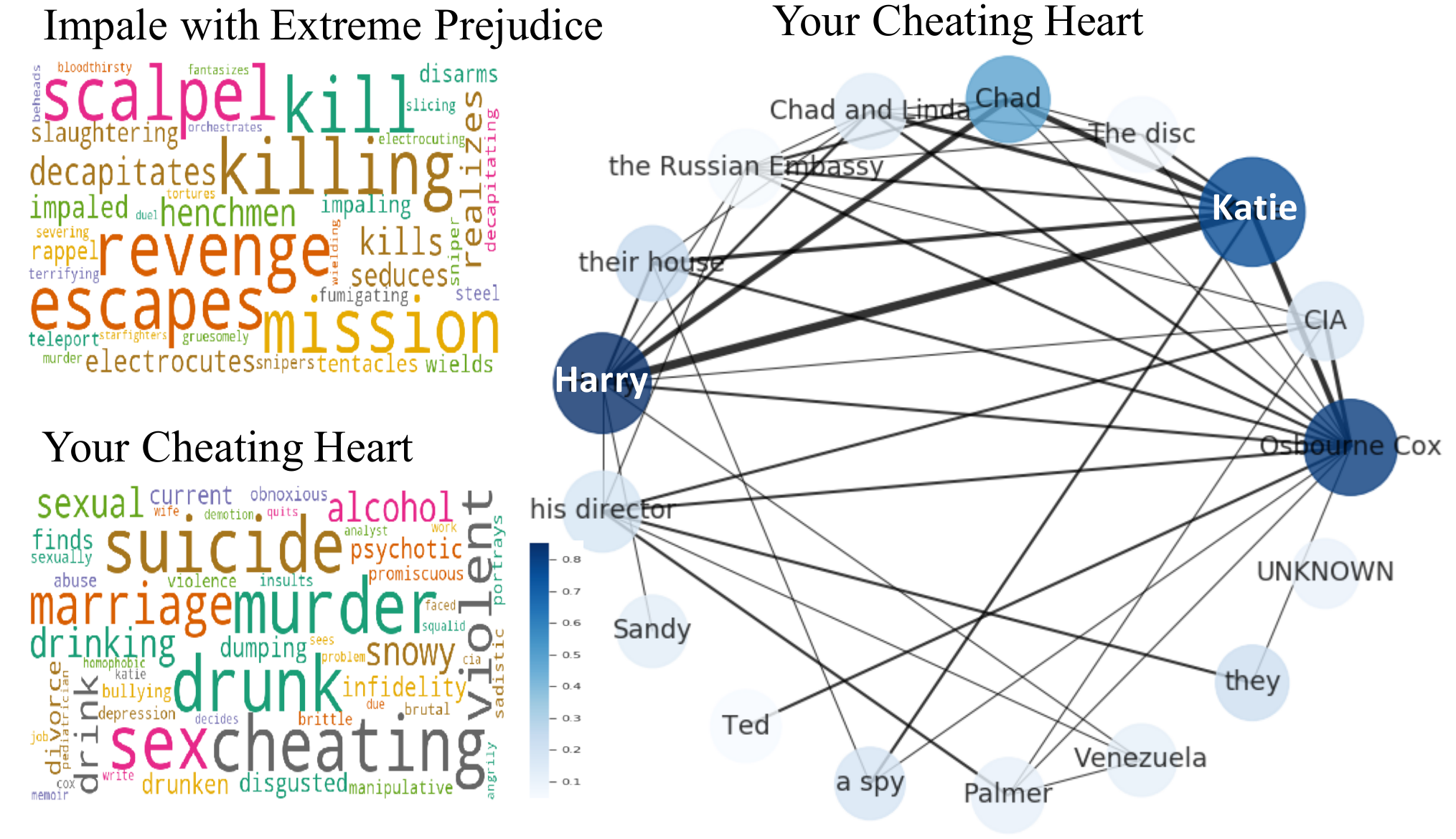}
  \caption{Word attention and graph attention. The left two word clouds display the most attended words by the Trope Embedding of ``Impale with Extreme Prejudice” and ``Your Cheating Heart”. The rightmost graph shows Trope Embedding of ``Your Cheating Heart”'s attention on the plot graph of the film \textit{Burn After Reading}. The attention focuses on ``Harry” who cheated the most in the film.
  We can see that even though Trope Embedding is randomly initialized, it successfully learns a meaningful representation for each trope and obtain meaningful attention. (cf. Section \ref{section:analysis:quality})}
  \label{fig:att}
\end{figure}

The Trope Embedding component in MulCom allows the attention mechanism to attend to input instances using tropes. 
We show some attention to the synopsis graph and word vector in Figure \ref{fig:att} to demonstrate the reasoning process when making predictions. The left word-clouds demonstrate the most attended words for tropes. We can observe that ``scalpel”, ``kill”, ``revenge,” and ``escape” are highlighted when detecting \textit{Impale with Extreme Prejudice}. Also, the model pays attention to ``suicide”, ``murder”, ``drunk,” and ``sexcheating” when capturing \textit{Your Cheating Heart}. The right graph shows the attention matrix of MSRRN, focusing on ``Harry”, who cheats in the film.

\section{Human Evaluation and Discussion}\label{humaneval}
This section performs a comprehensive human analysis to a subset of \ourdataset{} test set. We discuss the trope detection \textit{inference process} of human, evaluate the \textit{answerability} of our web-collected dataset, and provide some \textit{potential direction} of tackling our \ourdataset{}, or collecting similar dataset for deep cognition and comprehension tasks.

\subsection{Human Inference Process}\label{section:humaneval:humanprocess}
Intuitively, an educated human could determine if a trope occurs in a movie by matching facts in the movie and the trope definition. It is tricky that the tropes are mostly annotated according to the video instead of the movie synopses. While it might cause some information loss, we also discover that humans could conceive some tropes by \textit{commonsense} instead of facts. For instance, a scene of fighting with knives might involve a trope \textit{Off His Head}, though it is not exhibited due to being too detail. We see the potential to extend our work to involve commonsense related topics. Therefore, we consider conceivable tropes in our human evaluation and hope to point to future research.


\subsection{Setup}\label{section:humaneval:setup}
\begin{figure}
    \centering
    \includegraphics[width=\columnwidth]{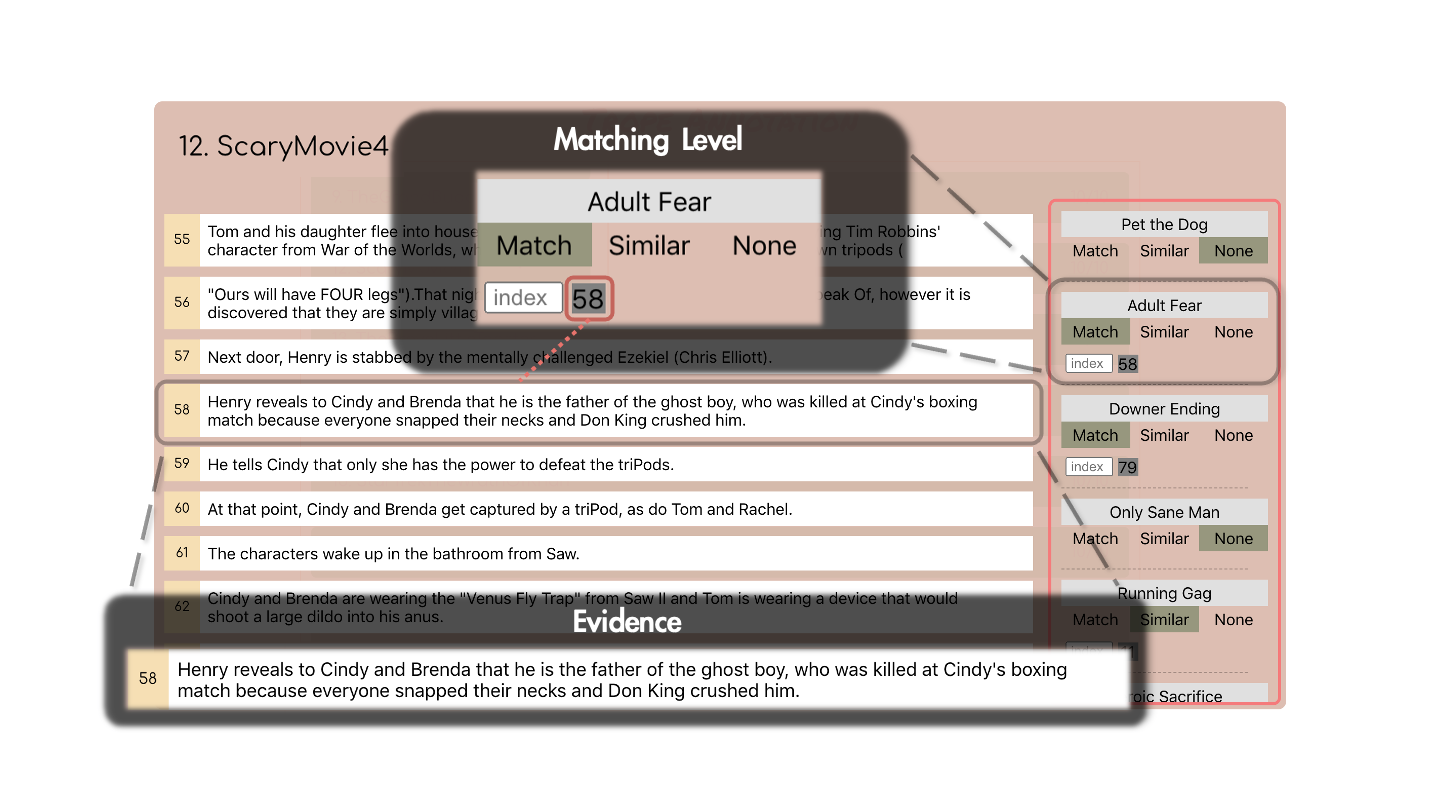}
    \caption{Our trope human evaluation interface. Annotators are instructed to select ``match” when there is clear evidence of a trope in the movie synapse and select ``similar” when a trope is not directly portrayed but conceivable. (cf. Section \ref{section:humaneval:humanprocess}) }
    \label{fig:tropeanno}
\end{figure}

We conduct human evaluation with 40 movie synopses in the test-set. For each movie, we sample 10 trope candidates, including positive and negative ones. We ask human annotators to choose ``match” when there is clear evidence of a trope and choose ``similar” if a trope is not shown but conceivable with the context and the commonsense, to choose ``none” if there are no cues of a trope. The human evaluation interface is shown in Figure \ref{fig:tropeanno}. 

\subsection{Human Performance}\label{section:humaneval:performance}



\begin{table}[ht]
    \centering
    \begin{tabular}{|c|c|c|c|}
        \hline
        Inference & Precision & Recall & F1 \\ \hline
        Machine (MulCom) & 35.98 & 22.23 & 27.48 \\ \hline
        Human (Fact only) & 56.25 & 44.63 & 49.77\\ \hline
        Human (Fact + Commonsense) & 65.77 & 63.98 & 64.87 \\ \hline
    \end{tabular}
    \caption{Human Performance. A considerable gap is shown between the best learning system and humans, suggesting that trope detection is challenging. (See setting in Section \ref{section:humaneval:setup} and discussions in Section \ref{section:humaneval:performance}}
    \label{tab:humanevalres}
\end{table}
We evaluate two different scenarios: (1) Fact only where only \textit{match} is considered true, (2) Fact + Commonsense where both \textit{match} and \textit{similar} are considered as true. 

As shown in Table \ref{tab:humanevalres}, human performance surpasses best machine performance \footnote{We evaluate sampled movie synopses and tropes only} in our experiments in a large margin in both precision and recall, resulting in 1.81 times of F1 score only considering match, and 2.93 times of F1 score when involving common sense.

Comparing fact only (second row) and fact+commonsense (third row), involving commonsense improves not only recall but also precision and results in a much better F1 score (15.1 F1 gain), and suggests that commonsense is crucial prior knowledge for trope detection.

\subsection{Potential Directions}
\paragraph{Commonsense} Human performance has shown that commonsense boost human performance by 15 points of F1. From a learning system aspect, utilizing knowledge that might not be exhibited in a movie might be a direction worth considering. 

\paragraph{Video Trope Detection} Visual cognition is another emerging topic. Future research could use our trope detection framework to annotate tropes on movies or TV series. It would also be interesting to investigate the machine capability of video trope detection, compared to existing video comprehension tasks such as Movie Question Answering \cite{MovieQA,pororoqa_ijcai2017-280,tvqa_lei-etal-2018-tvqa} or Video Language Inference \cite{violin_Liu_2020_CVPR}.

\section{Conclusion}
In this work, we presented a brand new task with a new dataset \ourdataset{} for situation and behavior understanding in films. Unlike previous tasks and datasets, trope detection requires deep cognitive skills, including consciousness, systematic generalization, and casual and motivational comprehension. Modern learning systems including contextual embedding BERT, movie tag prediction system, and recurrent relational network reach at most 23.97 F1 score, which is far behind human performance (64.87) and suggests that trope detection is a challenging task. We proposed a Multi-Level Comprehension framework with multi-stream attention and multi-step reasoning to tackle this problem and boost performance to 25.00 F1. We carefully analyze the task along with the network's behavior. We adopt human evaluation to verify the answerability of our dataset and discuss potential directions. We believe that our research could pave a new path for future research on deep cognition. 

\section*{Acknowledgement}
This work was supported in part by the Ministry of Science and Technology, Taiwan, under Grant MOST 109-2634-F-002-032. We benefit from NVIDIA DGX-1 AI Supercomputer and are grateful to the National Center for High-performance Computing. We thank Dr. John R. Smith, IBM T. J. Watson Research Center, New York, for his sharing in film tropes. We also thank Igor Morawski and Yun-Hsuan Liu for participating the human evaluation. 

\bibliographystyle{ACM-Reference-Format}
\bibliography{main}
\end{document}